%% file: main.tex
\documentclass{article}

\usepackage{microtype}
\usepackage{graphicx}
\usepackage{dblfloatfix}
\usepackage{subfigure}
\usepackage{tablefootnote}
\usepackage{booktabs} 

\input{math_commands.tex}

\usepackage{xcolor}
\usepackage{amsfonts}
\usepackage{booktabs}
\usepackage{hyperref}
\usepackage{url}

\usepackage{hyperref}

\usepackage[accepted]{icml2023}

\usepackage{amsmath}
\usepackage{amssymb}
\usepackage{mathtools}
\usepackage{amsthm}

\usepackage[capitalize,noabbrev]{cleveref}

\theoremstyle{plain}
\newtheorem{theorem}{Theorem}[section]

\theoremstyle{definition}
\newtheorem{definition}[theorem]{Definition}

\theoremstyle{remark}

\usepackage[textsize=tiny]{todonotes}

\icmltitlerunning{TAN Without a Burn: Scaling Laws of DP-SGD}

\newcommand{\pierre}[1]{}
\newcommand{\alex}[1]{}
\newcommand{\tom}[1]{}

\newcommand{\calM}{\mathcal{M}}
\newcommand{\defeq}{:=}
\newcommand{\epstan}{\varepsilon_{\mathrm{TAN}}}
\newcommand{\gauss}{\mathcal{N}}

\newcommand{\prob}{\mathbb{P}}

\DeclarePairedDelimiterX{\infdivx}[2]{(}{)}{%
  #1\;\delimsize\|\;#2%
}
\newcommand{\rdp}{D_\alpha\infdivx}

\begin{document}

\twocolumn[
\icmltitle{TAN Without a Burn: Scaling Laws of DP-SGD}




\begin{icmlauthorlist}
\icmlauthor{Tom Sander}{polytechnique,meta}
\icmlauthor{Pierre Stock}{meta}
\icmlauthor{Alexandre Sablayrolles}{meta}
\end{icmlauthorlist}

\icmlaffiliation{polytechnique}{CMAP, École polytechnique, Palaiseau, France}
\icmlaffiliation{meta}{Meta AI, Paris, France}

\icmlcorrespondingauthor{Tom Sander}{tomsander@meta.com}

\icmlkeywords{Machine Learning, ICML}

\vskip 0.3in
]



\printAffiliationsAndNotice{}  
\newcommand{\BigSigma}{TAN\@\xspace}
\begin{abstract}
Differentially Private methods for training Deep Neural Networks (DNNs) have progressed recently, in particular with the use of massive batches and aggregated data augmentations for a large number of training steps.
These techniques require much more computing resources than their non-private counterparts, shifting the traditional privacy-accuracy trade-off to a privacy-accuracy-compute trade-off and making hyper-parameter search virtually impossible for realistic scenarios. 
In this work, we decouple privacy analysis and experimental behavior of noisy training to explore the trade-off with minimal computational requirements.
We first use the tools of Rényi Differential Privacy (RDP) to highlight that the privacy budget, when not overcharged, only depends on the total amount of noise (TAN) injected throughout training.
We then derive scaling laws for training models with DP-SGD to optimize hyper-parameters with more than a $100\times$ reduction in computational budget.
We apply the proposed method on CIFAR-10 and ImageNet and, in particular, strongly improve the state-of-the-art on ImageNet with a $+9$ points gain in top-1 accuracy for a privacy budget $\varepsilon=8$.
\end{abstract}

\input{introduction}

\input{TANIcml2023/related.tex}

\input{TANIcml2023/method.tex}

\input{TANIcml2023/experiments.tex}

\input{TANIcml2023/conclusion.tex}

\nocite{langley00}

\bibliography{main}
\bibliographystyle{icml2023}

\newpage
\appendix
\onecolumn
\input{appendix}

\end{document}

%% file: math_commands.tex
\usepackage{amsmath,amsfonts,bm,xspace}

\def\eqref#1{equation~\ref{#1}}

\def\1{\bm{1}}

\DeclareMathAlphabet{\mathsfit}{\encodingdefault}{\sfdefault}{m}{sl}
\SetMathAlphabet{\mathsfit}{bold}{\encodingdefault}{\sfdefault}{bx}{n}

\newcommand{\E}{\mathbb{E}}



%% file: introduction.tex

\vspace{-1em}
\section{Introduction}
\label{Intro}

Deep neural networks (DNNs) have become a fundamental tool of modern artificial intelligence, producing cutting-edge performance in many domains such as computer vision~\citep{ResNets}, natural language processing~\citep{devlin2018bert} or speech recognition \citep{amodei2016deep}.
The performance of these models generally increases with their training data size~\citep{GPT, gopher, Dalle-2}, which encourages the inclusion of more data in the model's training set.
This phenomenon also introduces a potential privacy risk for data that gets incorporated. 
Indeed, AI models not only learn about general statistics or trends of their training data distribution (such as grammar for language models), but also remember verbatim information about individual points (e.g., credit card numbers), which compromises their privacy~\citep{carlini2019secret, carlini2021extracting}.
Access to a trained model thus potentially leaks information about its training data.

\begin{figure}[t]
    \centering
    \includegraphics[width=\columnwidth]{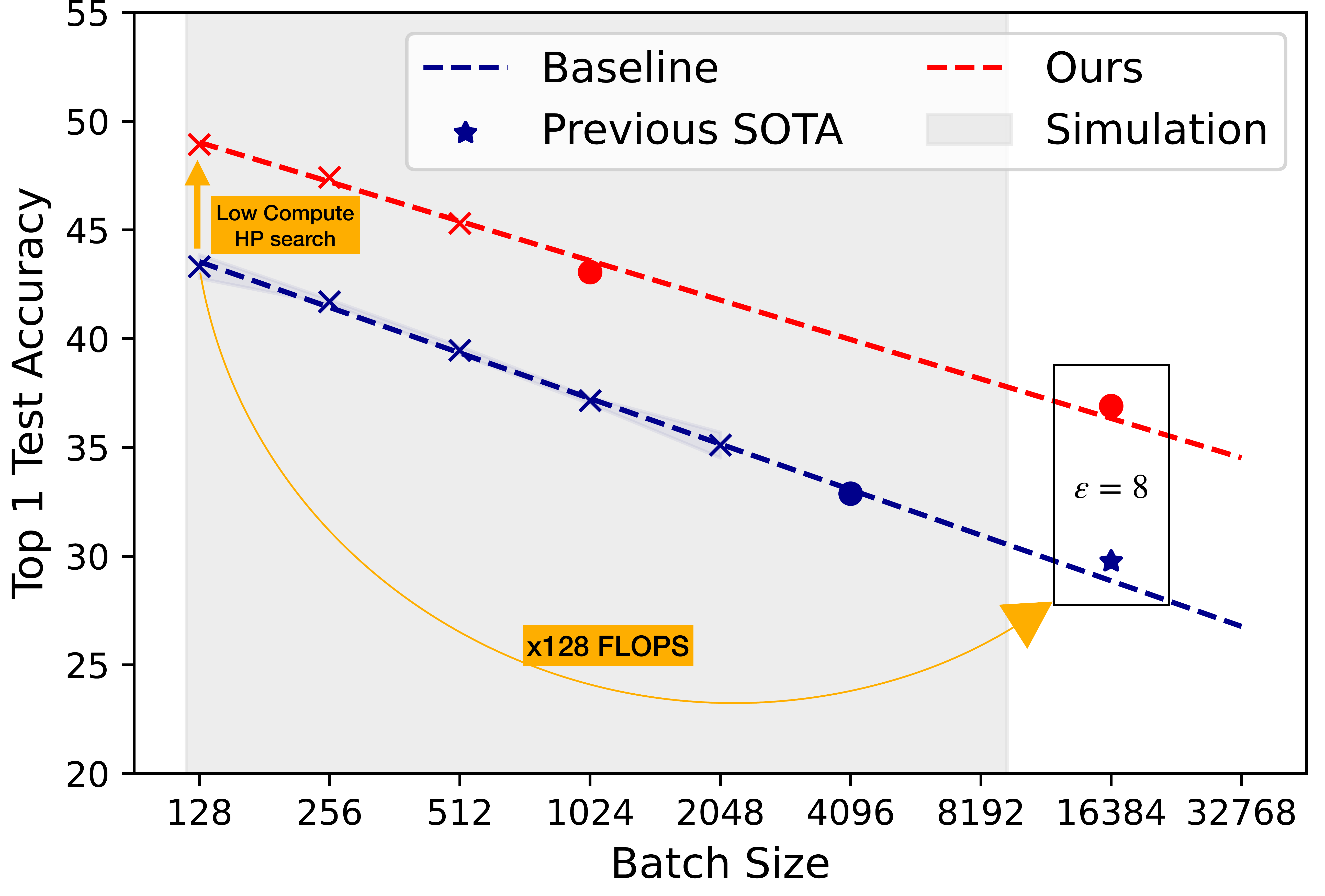}
    \vspace{-1em}
    \caption{
    Training from scratch with DP-SGD on ImageNet. All points are obtained at constant number of steps $S=72k$ and constant ratio $\sigma/B$, with $\sigma_{\mathrm{ref}}=2.5$ and $B_{\mathrm{ref}}=16384$. 
 The dashed lines are computed using a linear regression on the crosses, and the dots and stars illustrate the predictive power of TAN. We perform low compute hyper-parameter (HP) search at batch size $128$ and extrapolate our best setup for a single run at large batch size: stars show our reproduction of the previous SOTA from \cite{DeepMindUnlocking} and improved performance obtained under the privacy budget $\varepsilon=8$ with a $+6$ points gain in top-1 accuracy. The shaded blue areas denote 2 standard deviations over three runs.
 \vspace{-1em}
 }
    \label{fig:Figure1}
\end{figure}

The gold standard of disclosure control for individual information is Differential Privacy (DP)~\citep{dwork2006calibrating}.
Informally, DP ensures that the training does not produce very different models if a sample is added or removed from the dataset. 
Motivated by applications in deep learning, DP-SGD~\citep{abadi2016deep} is an adaptation of Stochastic Gradient Descent (SGD) that clips individual gradients and adds Gaussian noise to their sum. 
Its DP guarantees depend on the privacy parameters: the sampling rate $q=B/N$ (where $B$ is the batch size and $N$ is the number of training samples), the number of gradient steps $S$, and the noise $\sigma^2$.

\looseness=-1 Training neural networks with DP-SGD has seen progress recently, due to several factors.
The first is the use of pre-trained models, with DP finetuning on downstream tasks~\citep{li2021large,DeepMindUnlocking}.
This circumvents the traditional limitations of DP, because the model learns meaningful features from public data and can adapt to downstream data with minimal information.
In the remainder of this paper, we only consider models trained \emph{from scratch}, as we focus on obtaining information through the DP channel.
Another emerging trend among DP practitioners is to use massive batch sizes at a large number of steps to achieve a better tradeoff between privacy and utility: \citet{DPBert} have successfully pre-trained BERT with DP-SGD using batch sizes of $2$ million. 
This paradigm makes training models computationally intensive and hyper-parameter (HP) search effectively impractical for realistic datasets and architectures.

\looseness=-1 In this context, we look at DP-SGD through the lens of the Total Amount of Noise (TAN) injected during training, and use it to decouple two aspects: privacy accounting and influence of noisy updates on the training dynamics.
We first observe a heuristic rule: when typically $\sigma >2$
, the privacy budget $\varepsilon$ only depends on the total amount of noise.

Using the tools of RDP accounting, we approximate $\varepsilon$ by a simple closed-form expression.
We then analyze the scaling laws of DNNs at constant TAN and show that performance at very large batch sizes (computationally intensive) is predictable from performance at small batch sizes as illustrated in Figure~\ref{fig:Figure1}. 
Our contributions are the following:

\begin{itemize}
    \item We take a heuristic view of privacy accounting by introducing the Total Amount of Noise (TAN) and show that in a regime when the budget $\varepsilon$ is not overcharged, it only depends on TAN;
    \item We use this result in practice and derive scaling laws that showcase the predictive power of TAN to reduce the computational cost of hyper-parameter tuning with DP-SGD, saving a factor of $128$ in compute on ImageNet experiments (Figure~\ref{fig:Figure1}). 
    We then use TAN to find optimal privacy parameters, leading to a gain of $+9$ points under $\varepsilon=8$ compared to the previous SOTA;
    \item We leverage TAN to quantify the impact of the dataset size on the privacy/utility trade-off and show that with well chosen privacy parameters, doubling dataset size halves $\varepsilon$ while providing better performance. 
\end{itemize}

%% file: TANIcml2023/related.tex
\section{Background and Related Work}
\label{sec:related}

\begin{figure*}[!t]
    \centering
    \includegraphics[width=0.95\textwidth]{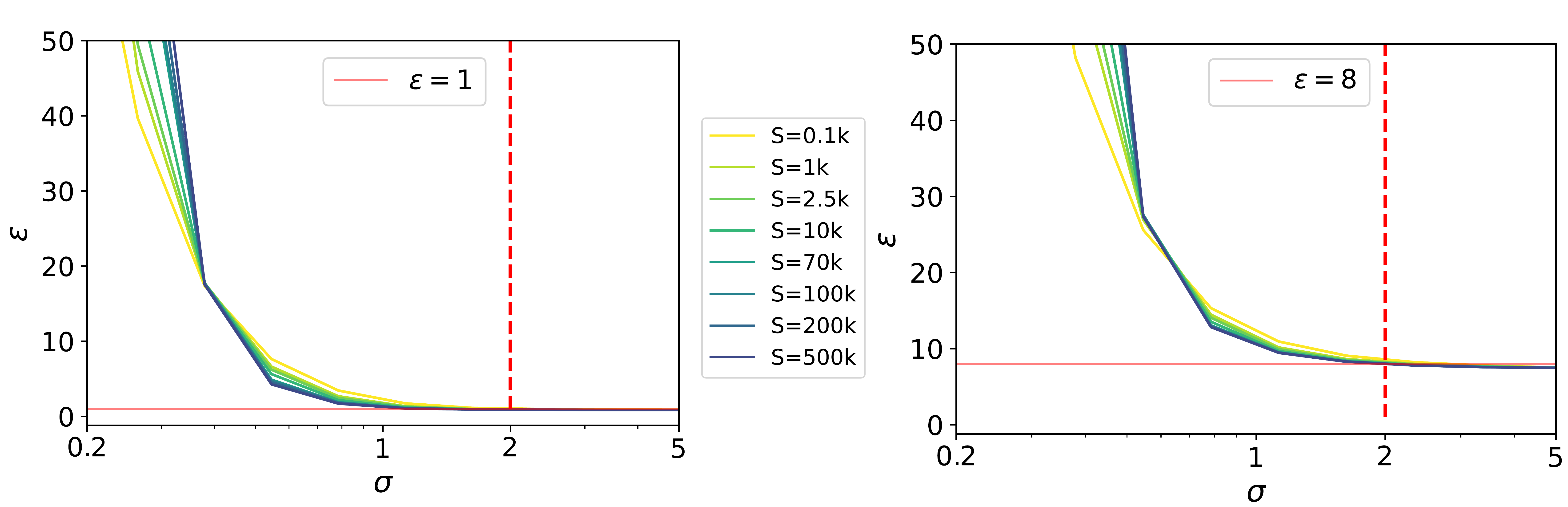}
    \vspace{-3mm}
    \caption{
    Privacy budget $\varepsilon$ as a function of the noise level $\sigma$ with $\eta$ constant.
    On both figures, each curve corresponds to a different number of steps $S$, and each point on the curve is computed at a sampling rate $q$ such that $\eta$ is constant. 
    On the left, we use $\eta=0.13$ (resulting in $\varepsilon_{\mathrm{TAN}}=1$ in Equation~\ref{eq:eps_approx_BigS}). On the right, we use $\eta=0.95$ ($\varepsilon_{\mathrm{TAN}}=8$). We observe a ``privacy wall" imposing $\sigma \geq 0.5$ for meaningful level of privacy budget $\varepsilon$, and $\sigma \geq 2$ for constant $\varepsilon \approx \varepsilon_{\mathrm{TAN}}$.}
    \vspace{-3mm}
    \label{fig:privacy_budget}
\end{figure*}

In this section, we review traditional definitions of DP, including Rényi Differential Privacy.
We consider a randomized mechanism $\calM$ that takes as input a dataset $D$ of size $N$ and outputs a machine learning model $\theta \sim \calM(D)$.
\begin{definition}[Approximate Differential Privacy]
A randomized mechanism $\calM$ satisfies $(\varepsilon, \delta)$-DP~\citep{dwork2006calibrating} if, for any pair of datasets $D$ and $D'$ that differ by one sample and for all subset $R\subset \mathbf{Im}(\calM)$,
\begin{equation}
    \prob(\calM(D) \in R) \leq \prob(\calM(D') \in R) \exp( \varepsilon) + \delta.
\end{equation}
\end{definition}

DP-SGD~\citep{abadi2016deep} is the most popular DP algorithm to train DNNs.
It selects samples uniformly at random with probability $q=B/N$ (with $B$ the batch size and $N$ the number of training samples), clips per-sample gradients to a norm $C$ ($\mathrm{clip}_C$), aggregates them and adds (gaussian) noise. With $\theta$ the parameters of the DNN and $\ell_i(\theta)$ the loss evaluated at sample $(x_i,y_i)$, it uses noisy gradient 
\begin{align}\label{eq:DP_SGD}
g := \frac{1}{B}\sum_{i \in B} \text{clip}_C\left(\nabla_\theta \ell_i(\theta)\right) + \gauss \left(0,  \frac{C^2 \sigma^2}{B^2} \right)
\end{align}

to train the model. The traditional privacy analysis of DP-SGD is obtained through Rényi Differential Privacy.
\begin{definition}[Rényi Divergence]
    For two probability distributions $P$ and $Q$  defined over $\mathcal R$, the Rényi divergence of order $\alpha > 1$ of $P$ given $Q$ is: 
    \begin{equation*}
        \rdp{P}{Q} \defeq \frac{1}{\alpha - 1}\log \E_{x\sim Q}\left(\frac{P(x)}{Q(x)}\right)^\alpha.
    \end{equation*}
\end{definition}

\begin{definition}[Rényi DP]
    A randomized mechanism $\calM \colon \mathcal D \to \mathcal R$ satisfies $(\alpha, d_\alpha)$-Rényi differential privacy (RDP) if, for any $D, D\in \mathcal D'$ that differ by one sample:
    \begin{equation*}
        \rdp{\calM(D)}{\calM(D')} \leq d_\alpha.
    \end{equation*}
\end{definition}

RDP is a convenient notion to track privacy because composition is additive: a sequence of two algorithms satisfying $(\alpha, d_\alpha)$ and $(\alpha, d'_\alpha)$ RDP satisfies $(\alpha, d_\alpha+d'_\alpha)$ RDP.
In particular, $S$ steps of a $(\alpha, d_\alpha)$ RDP mechanism satisfiy $(\alpha, S d_\alpha)$ RDP.
\citet{mironov2019r} show that each step of DP-SGD satisfies $(\alpha, g_\alpha(\sigma,q))$-RDP with 
\begin{equation*}
    g_\alpha(\sigma,q) \defeq \rdp{(1-q)\gauss(0,\sigma^2)+q\gauss(1,\sigma^2)}{\gauss(0,\sigma^2)}.
\end{equation*}

Finally, a mechanism satisfying $(\alpha, d_\alpha)$-RDP also satisfies $(\varepsilon, \delta)$-DP~\citep{mironov2017renyi}  for $\varepsilon = d_\alpha + \frac{\log(1/\delta)}{\alpha-1}$.
Performing $S$ steps of DP-SGD satisfies  $(\varepsilon_{RDP}, \delta)$-DP with
\begin{align}
\label{eq:eps_rdp}
\varepsilon_\mathrm{RDP} \defeq \min_\alpha S g_\alpha(\sigma,q) + \frac{\log(1/\delta)}{\alpha-1}.
\end{align}
RDP is the traditional tool used to analyse DP-SGD, but other accounting tools have been proposed to obtain tighter bounds~\citep{gopi2021numerical}. 
In this work, we use the accountant due to \cite{balle2020hypothesis}, whose output is referred to as $\varepsilon$, which is slightly smaller than $\varepsilon_\mathrm{RDP}$.



\paragraph{DP variants}
Concentrated Differential Privacy (CDP) \citep{dwork2016concentrated,bun2016concentrated} was originally proposed as a relaxation of $(\varepsilon,\delta)$- DP with better compositional properties. 
Truncated CDP (tCDP) \citep{bun2018composable} is an extension of CDP, with improved properties of privacy amplification via sub-sampling, which is crucial for DP-SGD-style algorithms. 
The canonical noise for tCDP follows a ``sinh-normal"' distribution, with tails exponentially tighter than a Gaussian. 
In Sections \ref{sec:RDP} and \ref{sec:connection}, we highlight the practical implications of the Privacy amplification by sub-sampling behavior of DP-SGD. 
We observe that in the large noise regime, $\varepsilon_{\text{RDP}}$ can be approximated by a very simple closed form expression of the parameters $(q,S,\sigma)$ through \BigSigma, and relate it to CDP and tCDP. 

 \vspace{-1em}
\paragraph{Training from Scratch with DP.}
Training ML models with DP-SGD typically incurs a loss of model utility, but using very large batch sizes improves the privacy/utility trade-off ~\citep{DPBert, li2021large}. 
\citet{DeepMindUnlocking} recently introduced Augmentation Multiplicity (AugMult), which averages the gradients from different augmented versions of every sample before clipping and leads to improved performance on CIFAR-10. 
Computing per-sample gradients with mega batch sizes for a large number of steps and AugMult makes DP-SGD much more computationally intensive than non-private training, typically dozens of times. For instance, reproducing the previous SOTA on ImageNet of \citet{DeepMindUnlocking} under $\varepsilon=8$ necessitates a $4$-day run using $32$ A100 GPUs, while the non-private SOTA can be reproduced in a few hours with the same hardware \citep{goyal2017accurate}. \citet{LowRank} propose to use low-rank reparametrization of the weight matrices to diminish the computational cost of accessing per-sample gradients.

\paragraph{Finetuning with DP-SGD.}
\citet{tramer2020differentially} show that handcrafted features are very competitive when training from scratch, but fine-tuning deep models outperforms them.
\citet{li2021large,yu2021differentially} fine-tune language models to competitive accuracy on several NLP tasks.
\citet{DeepMindUnlocking} consider models pre-trained on JFT-300M and transferred to downstream tasks.

%% file: TANIcml2023/method.tex
\section{The \BigSigma approach}

\begin{figure*}[!ht]
    \centering
    \includegraphics[width=0.99\textwidth]{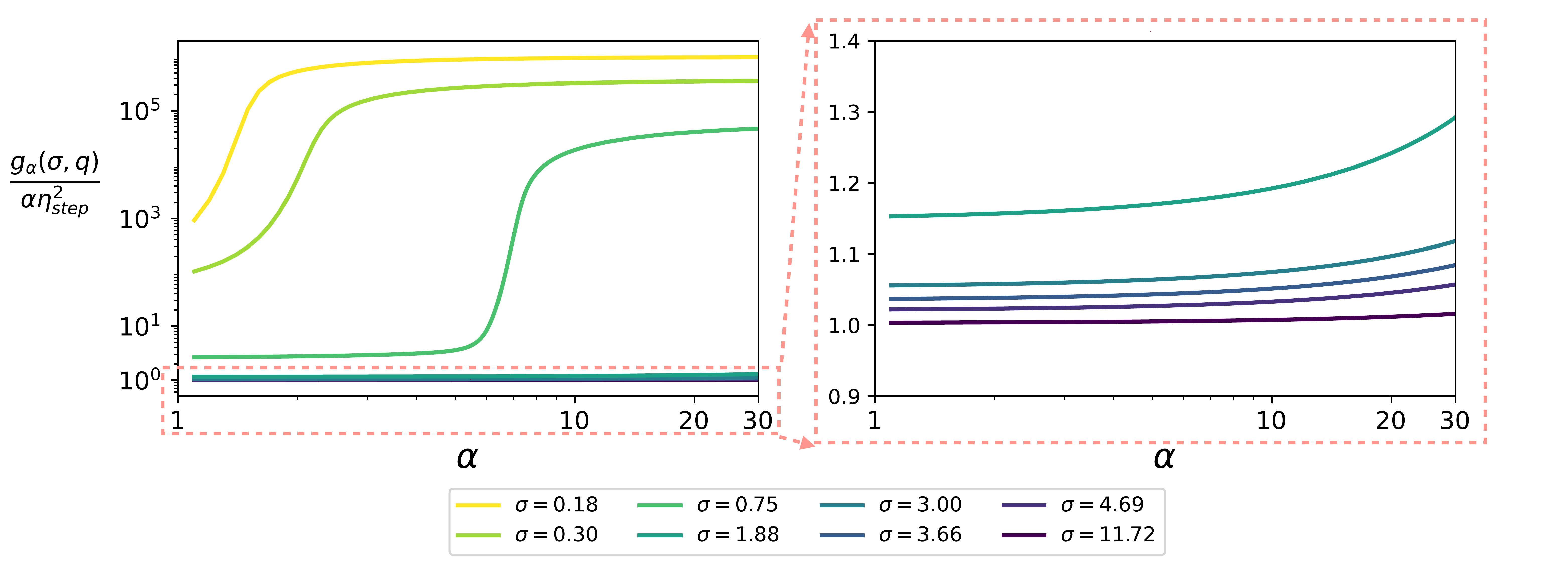}
    \caption{
    Approximation of $g_{\alpha}(\sigma,q)$. All curves correspond to distinct couples $(q,\sigma)$ such that $\eta_\mathrm{step}=3.9\times 10^{-3}$ (used for ImageNet). The right plot corresponds to an enlargement of the left plot: the ratio is very close to $1$ for $\sigma\geq2$. \label{fig:transition_phase}}
\end{figure*}

We introduce the notion of Total Amount of Noise (\BigSigma) and discuss its connections to DP accounting. 
We then demonstrate how training with reference privacy parameters $(q_\mathrm{ref},\sigma_\mathrm{ref},S)$ can be simulated with much lower computational resources using the same TAN with smaller batches.

\begin{definition}
Let the individual signal-to-noise ratio $\eta$ (and its inverse $\Sigma$, the Total Amount of Noise or TAN) be:
\begin{equation*}
\eta^2 = \frac{1}{\Sigma^2} \defeq \frac{q^2 S} {2 \sigma^2}.
\end{equation*}
\end{definition}

\subsection{Motivation}\label{sec:motivation}


We begin with a simple case to motivate our definition of \BigSigma. 
We assume a one-dimensional model, where the gradients of all points are clipped to $C$. 
Looking at Equation~\ref{eq:DP_SGD}, in one batch of size $B$, the expected signal from each sample is $C/B$ with probability $q=B/N$ and $0$ otherwise.
Therefore, the expected individual signal of each sample after $S$ steps is $S C/N$, and its squared norm is $ S^2 C^2 / N^2$.
The noise at each step being drawn independently, the variance across $S$ steps adds to $ S C^2 \sigma^2 / B^2$. 
The ratio between the signal and noise is thus equal to (up to a factor $1/2$)

\begin{align*}
\frac{\frac{S^2 C^2}{N^2}}{\frac{2 S C^2 \sigma^2}{B^2}} = \frac{q^2 S} {2 \sigma^2} = \eta^2.
\end{align*}

\looseness=-1 Denoting $\eta_\mathrm{step} := q/\sqrt{2}\sigma$, we have $\eta^2 = S\eta_{\mathrm{ step}}^2$. 
The ratio $\sigma/q$ is noted by \citet{li2021large} as the effective noise. 
The authors found that for a fixed budget $\varepsilon$ and fixed $S$, the effective noise decreases with $B$. 
Our analysis goes further by analyzing how RDP accounting explains this dependency.

\vspace{-1mm}
\subsection{Connection with Privacy Accounting}
\label{sec:RDP}

Intuitively, we expect that the privacy budget $\varepsilon$ only depends on the signal-to-noise ratio $\eta$ (an approximation of the extracted information). 
In Figure~\ref{fig:privacy_budget}, we plot $\varepsilon$ as a function of $\sigma$ and $S$, at \emph{fixed} $\eta$, and observe that $\varepsilon$ is indeed constant, but only when $\sigma > 2$.
When $\sigma$ gets smaller, $\varepsilon$ surges, creating a ``Privacy Wall".
We shed light on this phenomenon by looking at the underlying RDP values. We observe in Figure~\ref{fig:transition_phase} that when $\sigma > 2$, $g_{\alpha}(\sigma,q)$ is close to $\alpha q^2 / (2\sigma^2)= \alpha\eta_{\mathrm{step}}^2$. Conversely, when $\sigma < 2$, $g_\alpha$ becomes much larger than $\alpha \eta_\mathrm{step}^2$, which explains the blow-up in $\varepsilon$ from Figure~\ref{fig:privacy_budget}.

The existence of a phase transition with a sub-sampled Gaussian mechanism is also noticed in \citet{abadi2016deep, mironov2019r}. \citet{wang2019subsampled} identify that it typically happens when $q\alpha\exp{(\alpha/2\sigma^2)} > 1$ for RDP, implying that $g_{\alpha}(\sigma,q)=O(\alpha q^2/\sigma^2)$ for large $\sigma$. We deliberately dispose of the big $O$ notation and inject our refined (empirical) approximation $g_{\alpha}(\sigma,q)\approx\alpha\eta_{\mathrm{step}}^2$ in the definition of $\varepsilon_{\mathrm{RDP}}$ (\eqref{eq:eps_rdp}). We get: 
\begin{equation}
\begin{split}
\label{eq:eps_approx_BigS}
    \varepsilon_{\mathrm{RDP}} &\approx \eta^2 + \min_{\alpha}\left((\alpha-1)\eta^2 + \frac{\log\left(1/\delta\right)}{\alpha-1}\right) \\ &= \eta^2 + 2\eta\sqrt{\log\left(1/\delta\right)} =: \epstan(\eta).
\end{split}
\end{equation}

\begin{figure*}[!ht]
    \centering
    \includegraphics[width=0.49\textwidth]{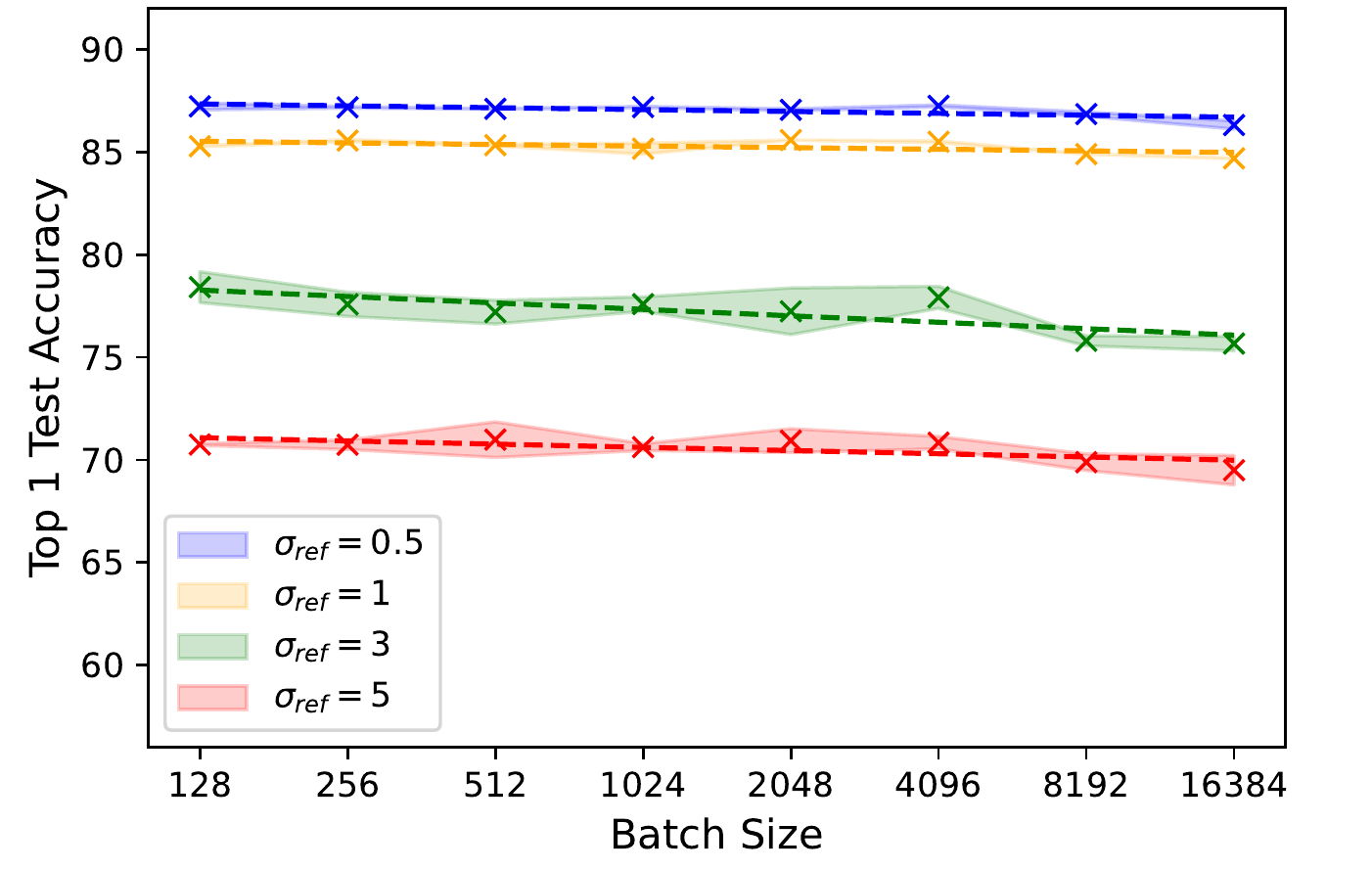}
    \includegraphics[width=0.49\textwidth]{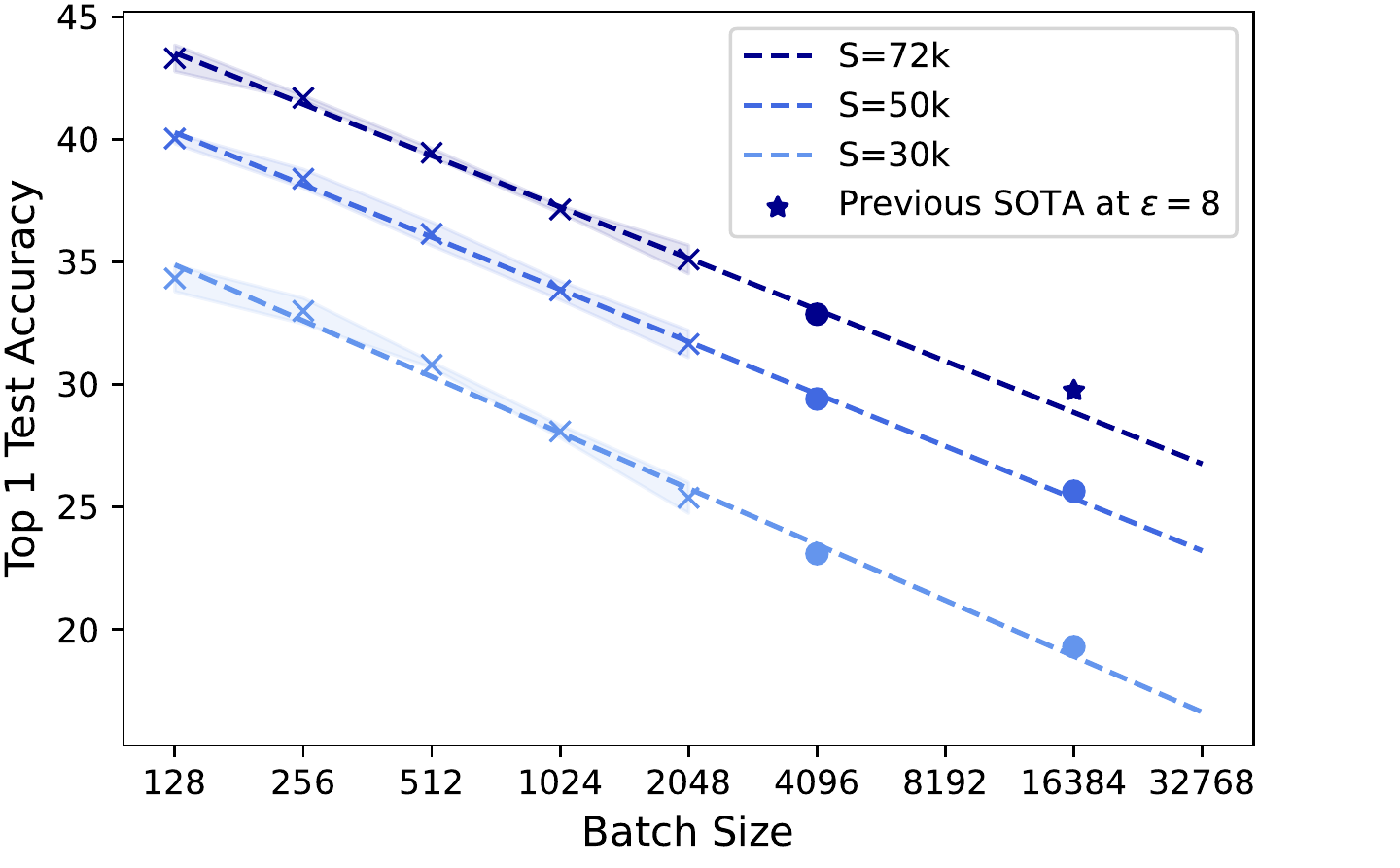}
    \caption{
    \label{fig:bigsigma_cifar}
    Test accuracies at constant $\eta_\mathrm{step}=B_{ref}/(\sqrt{2}N\sigma_{ref})$ and $S$ are (log) linearly decreasing with $B$.
    Dashed lines are computed using a linear regression on the crosses. 
    Shaded areas correspond to $3$ std confidence intervals. (Left) 
    CIFAR-10 with 16-4-WideResNet for $S_{ref}=2500$ steps and $B_\text{ref}=4096$. 
    Each curve corresponds to a different value of $\eta_\mathrm{step}$. 
    (Right) ImageNet with NF-ResNet-50 with various numbers of steps, all with $\sigma_{\text{ref}}=2.5$ and $B_\mathrm{ref}=16384$. 
    The scaling law holds for various training configurations.}
\end{figure*}

We verify this relationship empirically, and in particular choose $\eta$ to get a desired $\varepsilon_{\mathrm{TAN}}$ in Figure~\ref{fig:privacy_budget}. Having this simple approximation for $\varepsilon$ is useful because it allows for easy mental gymnastics:
for instance, doubling the sampling rate $q$ while dividing the number of steps $S$ by $4$ should leave the privacy budget constant, which we observe empirically. 
We suggest to leverage $\varepsilon_{\text{TAN}}$ as an approximation of the privacy budget that enables quick mental operations. To report the actual $\varepsilon$ accurately, we resort to the traditional RDP accounting method from \citet{balle2020hypothesis}.

\vspace{-0.5em}
\subsection{Connection to other notions of DP}
\label{sec:connection}
\citet{bu2020deep} show that under the assumptions of the Central Limit Theorem, the Gaussian Differential Privacy (GDP) parameter of DP-SGD is $q\sqrt{S(\exp{(1/\sigma^2)-1})}$. 
For $\sigma$ large, $\exp{(1/\sigma^2)-1} \approx 1/\sigma^2$, which means that the GDP parameter also becomes a function of \BigSigma only. 
We also note that if DP-SGD were $\eta^2$-CDP, the translation to $( \varepsilon, \delta)$- DP of \citet{bun2016concentrated} (Proposition 1.3) would be the same as $\varepsilon_{\text{ TAN }}$. 
The tCDP definition was chosen to better account for privacy amplification by sub-sampling (see Section \ref{sec:related}), typically to avoid the kind of exploding behaviour described in Section \ref{sec:RDP}. Our observation suggests that in the large noise regime that we are considering ($\sigma > 2$), $S$ steps of DP-SGD are approximately $(\eta^2,\omega)$-tCDP for $\omega$ such that $\log{(1/\delta) \leq (\omega-1)^2 \eta^2}$ (see Lemma 6 in \citet{bun2018composable}). 
Our approach differs because we observe this relationship as an empirical phenomenon and propose a simple heuristic criterion ($\sigma>2$) for the validity of our approximation. 
The (approximate) reduction of privacy accounting to $\eta^2=q^2 S/2 \sigma^2$ implies various ways to change $(q, \sigma, S)$ at a constant privacy budget.

\subsection{Scaling at Constant \BigSigma}
\label{sec:scaling}

Starting from $(q,\sigma,S)$, while $\sigma < 2$, we can double both $q$ and $\sigma$. 
It will drastically improve privacy accounting (Figure~\ref{fig:privacy_budget}).
On the other hand, we expect that keeping constant $S$ and the per step signal-to-noise ratio $\eta_{\text{step}}$ should yield to similar performance, as detailed in Section \ref{sec:discussion}.
However, since $S$ is fixed, doubling $q$ doubles the computational cost. 

\vspace{-1em}
\paragraph{Batch Scaling Laws.}
We now analyse how this strategy affects the performance of the network.
In Figure~\ref{fig:bigsigma_cifar}, we perform this analysis on CIFAR-10 and ImageNet.
We find that for triplets $(q, \sigma, S)$ for which $q/\sigma=q_{\mathrm{ref}}/\sigma_{\mathrm{ref}}$ (keeping $\eta_\mathrm{step}$ constant), the performance of the network is almost constant for various levels of noise on CIFAR-10, and (log) linearly decreases with the batch size on ImageNet.
We discuss these observations further in Section \ref{sec:discussion}.

\paragraph{Choice of $\sigma$.}
If $\sigma<2$, simultaneously doubling $q$ and $\sigma$ has a small or negligible negative impact on accuracy (Figure~\ref{fig:bigsigma_cifar}) but it can greatly reduce the privacy budget  (Figure~\ref{fig:privacy_budget}).
Reciprocally, halving $\sigma$ and $q$ is slightly beneficial or neutral to the performance (Figure~\ref{fig:bigsigma_cifar}), and if $\sigma > 4$,  it keeps the privacy guarantees \textit{almost} unchanged (Figure~\ref{fig:privacy_budget}). It also divides the computational cost by $2$.
This explains why state-of-the-art approaches heuristically find that mega-batches work well: a blind grid search on the batch size and the noise level at constant privacy budget is likely to discover batches large enough to have $\sigma > 2$.
Our analysis gives a principled explanation for the sweet spot of $\sigma \in [2,4]$ used by most SOTA approaches~\citep{DeepMindUnlocking,li2021large}.

\paragraph{Efficient \BigSigma Training.}
We further study the training process in the small batch size setting.
We choose the optimal hyper-parameters (including architecture, optimizer, type of data augmentation) in this simulated setting, and finally launch one single run at the reference (large) batch size, with desired privacy guarantees.
On ImageNet, simulating $B_\mathrm{ref}=16{,}384$ with $B=128$ thus reduces the computational requirements by a factor of 128.
Each hyper-parameter search for ImageNet at $B=16{,}384$ 
takes $4$ days using $32$ A100 GPUs; we reduce it to less than a day on a single A100 GPU.

%% file: TANIcml2023/experiments.tex
\section{Experiments}
\label{sec:experiments}

\setcounter{table}{1} 
\begin{table*}[b!]
\caption{\label{tab:hptuning}
Comparing optimal hyper-parameters. Keeping $\eta_{step}$ and $S$  constant, we compare various changes in the training pipeline. We compare with the baseline of \citet{DeepMindUnlocking} (blue line in Figure~\ref{fig:Figure1}: NFResNet-50, learning rate at 4, EMA decay at $0.99999$, $4$ random augmentations averaged over $3$ runs). Each gain is compared to the previous column.}
\vskip 0.15in
\begin{center}
\begin{small}
\begin{sc}
    \begin{tabular}{*{9}{c}}
    \toprule
    \multicolumn{9}{c}{Imagenet: $\sigma_\mathrm{ref}=2.5$, $B_\mathrm{ref}=16{,}384$, $S=72\mathrm{K}$} \\
    \midrule
    $B$ &  \multicolumn{2}{c}{$(lr,\mu,d)$} & \multicolumn{2}{c}{decay}  &  \multicolumn{2}{c}{AugMult} &  AugTest & Total\\ 
    \midrule
     128 & $(8,0,0)$ & +1.0 & 0.999 & +1.2 &  (Ours, 8) & +3.0 & +0.4 & +5.6\% \\
     256 & $(8,0,0)$ & +0.8 & 0.999 & +1.2 &  (Ours, 8) & +3.0 & +0.7 & +5.7\%\\
    512 & $(8,0,0)$ & +1.2 & 0.999 & +1.1 &  (Ours, 8) & +2.8 &  +1.1 &  +6.2\% \\
    1024 & $(8,0,0)$ & +1.6 & 0.999 & +1.2 &  (Ours, 8) & +2.3 & +0.8 & +5.9 \% \\
    \midrule
    16384 & - & - & - & - & - & - & +0.8 &\textbf{+6.7\%} \\
    \bottomrule 
\end{tabular}
\end{sc}
\end{small}
\end{center}
\vskip -0.1in
\end{table*}

We leverage our efficient \BigSigma training strategy and obtain new state-of-the-art results on ImageNet for $\varepsilon=8$ (Table~\ref{tab:SOTA}).
We then study the impact of the dataset size on the pricacy/utility trade-off. 
We also demonstrate how our low compute simulation framework can be used to detect performance bottlenecks when training with noisy updates: in our case, the importance of the order between activation and normalization in a WideResNet on CIFAR-10. 
\vspace{-1mm}
\subsection{Experimental Setup}

We use the CIFAR-10 dataset ~\citep{krizhevsky2009learning} which contains $50$K $32\times 32$   images grouped in $10$ classes. 
The ImageNet dataset~\citep{Imagenet,Imagenet2} contains $1.2$ million images partitioned into 1000 categories. 
For data augmentation, we always use Augmentation Multiplicity as detailed in Appendix ~\ref{appendix:augmult}.
For both datasets, we train models from random initialization. 
On CIFAR-10, we train 16-4-WideResNets \citep{WideResnets}. On Imagenet,  we compare Vision Transformers (ViTs) \citep{ViT}, Residual Neural Networks (ResNets) \citep{ResNets} and Normalizer-Free ResNets (NF-ResNets) \citep{NFnet}.
We always fix $\delta=1/N$ where $N$ is the number of samples and report the corresponding value of $\varepsilon$. 
We use $C=1$ for the clipping factor in Equation~\ref{eq:DP_SGD} as we did not see any improvement using other values. 
We use the Opacus~\citep{yousefpour2021opacus} and timm \citep{rw2019timm} libraries in Pytorch \citep{paszke2019pytorch}. We open-source the training code at \url{https://github.com/facebookresearch/tan}.

\vspace{-1mm}

\setcounter{table}{0} 
\begin{table}[t!]
\caption{
ImageNet top-1 test accuracy when training from scratch using DP-SGD. We use a NF-ResNet-50 with $\sigma=2.5$, hyper-parameters of Table~\ref{tab:hptuning} and  $(B,S)=(32768,18\mathrm{k})$ (Table~\ref{tab:varying_S_16384}). \textit{original} corresponds to the results stated in the paper, and \textit{reprod} to our reproduction of their results.}
\label{tab:SOTA}
\begin{center}
\begin{small}
\begin{sc}
    \begin{tabular}{lcc}
    \toprule
    Method &  $(\varepsilon,\delta)$ & Accuracy \\
    \midrule
     \cite{google6imagenet} &  $(13.2, 10^{-6})$ & 6.2\% \\
     \cite{DeepMindUnlocking} (\textit{original}) &  $(8, 8.10^{-7})$ & 32.4\% \\
     \cite{DeepMindUnlocking} (\textit{reprod}) &  $(8, 8.10^{-7})$ & 30.2\% \\
     Ours &  $(8, 8.10^{-7})$ & \textbf{39.2\%} \\
\bottomrule
\end{tabular}
\end{sc}
\end{small}
\end{center}
\vskip -0.1in
\end{table}

We decouple privacy hyper-parameters (HPs) from non-privacy HPs in our experiments. In Section \ref{sec:hp_tuning}, we use our simulated training with constant \BigSigma to find better non-privacy HPs at low compute keeping the privacy HPs ($B_{\text{ref}}, S, \sigma_{\text{ref}}$) fixed. 
In Section \ref{sec:varying_S}, we directly use \BigSigma to optimally choose better privacy HPs (which further improves performance by 3 points) and that constitutes our best state-of-the-art run (Table~\ref{tab:SOTA}).
We chose that baseline because the computational cost of each training run is high, thus corresponding to an ideal instantiation for our method.


\subsection{Hyper-parameter Tuning at Fixed \BigSigma}
\label{sec:hp_tuning}




We run a large hyper-parameter search and report the best hyper-parameters in Table~\ref{tab:hptuning} as well as the corresponding improvement for various batch sizes (at constant $\eta_\mathrm{step}$ and $S$). 
Each gain is compared to the optimal hyper-parameters find at the previous column.
We search over learning rates $\mathit{lr}\in[1,2,4,8,12,16]$, momentum parameters $\mu\in[0,0.1,0.5,0.9,1]$ and dampening factors $d\in[0,0.1,0.5,0.9,1]$.
We use exponential moving average (EMA) on the weights \citep{tan2019efficientnet} with a decay parameter in $[0.9,0.99,0.999,0.9999,0.99999]$.

We try different types of data augmentation, that we referred to as ``RRC", ``Ours" and ``SimCLR", and try for each various multiplicity of augmentations ($1, 2, 4, 8, 16$) (see Appendix~\ref{appendix:augmult} for details).
\begin{itemize}
    \item RRC: a standard random resized crop (crop chosen at random with an area between $8\%$ and $100\%$ of the original image and random aspect ratio in $[3/4, 4/3]$),
    \item Ours: random crop around the center with 20 pixels padding with reflect, random horizontal flip and jitter;
    \item SimCLR: the augmentation from \citet{simclr}, including color jitter, grayscale, gaussian blur and random resized crop, horizontal flip.
\end{itemize}


We find (Table~\ref{tab:hptuning}) that optimal parameters are the same in each scenario of simulation, as predicted in Section~\ref{sec:scaling}. We perform one run with these optimal parameters at $B=16384$ which satisfies a privacy budget of $\varepsilon=8$.
Note that we use multiple batch sizes only to support our hypothesis and batch scaling law, but it is sufficient to simulate only at $B=128$. 
Our experiments indicate that AugMult is the most beneficial when the corresponding image augmentations are rather mild.

\paragraph{Testing with Augmentations.}
\looseness=-1
We also test the model using a majority vote on the augmentations of each test image (AugTest column in Table~\ref{tab:hptuning}). 
We use the same type and number of augmentations as in training. 
It improves the final top-1 test accuracy.
This is in line with a recent line of work aiming at reconciling train and test modalities~\citep{touvron2019fixing}.
To provide a fair comparison with the state of the art, we decide \textbf{not to} include this gain in the final report in Table~\ref{tab:SOTA} and Table~\ref{tab:varying_S_16384}.
\paragraph{Choice of architecture and optimizer.} 
We have experimented with different architectures (ViTs, ResNets, NFResnets) and optimizers (DP-Adam, DP-AdamW, DP-SGD) (see Appendix~\ref{appendix:architecture} for details). 
Our best results are obtained using a NFResnet-50 and DP-SGD with constant learning rate and no momentum, which differs from standard practice in non-private training.

\subsection{Privacy Parameter Search at Fixed \BigSigma}
\label{sec:varying_S}

\setcounter{table}{2} 
\begin{table}[t]
\caption{Low compute simulation of privacy parameter search. We start from $B=256=16384/64$ and $S=72$K. We use $\sigma=2.5/64$ for all runs and no data augmentation.}
\label{tab:varying_S_256}
\vskip 0.15in
\begin{center}
\begin{small}
\begin{sc}
\begin{tabular}{*{4}{c}}
    \toprule
    \multicolumn{4}{c}{$B_\mathrm{ref}=256$, $S_\mathrm{ref}=72$K} \\
    \midrule
    $S$ & $B$ & $\mathit{lr}$ & Gain\\
    \midrule
     $9$K & 756 & 64 & -6.22\% \\
     $18$K & 512 & 32 & \textbf{+1.32\%} \\
     $72$K & 256 & 8 &  / \\
     $288$K & 128 & 2 & -1.88\% \\
\bottomrule
\end{tabular}
\end{sc}
\end{small}
\end{center}
\vskip -0.1in
\end{table}

\begin{table}[t]
    \centering
    \vspace*{-4mm}
    \caption{
    \label{tab:varying_S_16384}
    Privacy parameter search. We use the optimal parameters described in Section~\ref{sec:hp_tuning} with $\sigma=2.5$ for one expensive run and compare it with our optimal result}
    \vspace*{2mm}
    \begin{tabular}{*{5}{c}}
    \toprule
    \multicolumn{5}{c}{$B_\mathrm{ref}=16384$, $S_\mathrm{ref}=72$K} \\
    \midrule
    $S$ & $B$ & $\varepsilon$  &  $\mathit{lr}$ & Test acc\\
    \midrule
     $18$K & 32{,}768 & 8.00 &  32 & \textbf{39.2}\% \\
     $72$K & 16{,}384 & 7.97 &  8 & 36.1\% \\
\bottomrule
\end{tabular}
\end{table}
While we kept $S$ constant in previous experiments, we now explore constant TAN triplets $(q,\sigma,S)$ by varying $S$.
We keep $\sigma $ fixed to $2.5$ and vary $(B,S)$ starting from the reference $(16384,72\mathrm{K})$ at constant $\eta=q^2S/(2\sigma^2)$.
Given that $\sigma>2$, we stay in the \textit{almost} constant privacy regime (Figure~\ref{fig:privacy_budget}): we indeed observe $\varepsilon \approx \varepsilon_\mathrm{TAN}$ in Table~\ref{tab:varying_S_16384}. 
We scale the learning rate inversely to $S$ to compensate for the decrease of the noisy updates' magnitude (Equation~\ref{eq:DP_SGD}).
Since performing this privacy parameter search is computationally intensive, we first simulate training using our scaling strategy at $B=256$ (i.e. with the same $\eta_\mathrm{step}$) and display our results in Table~\ref{tab:varying_S_256}. 
Our best results are obtained for $18$k steps.
Finally, we perform one computationally expensive run at $S=18$k and $B=32768$, with other hyper-parameters from Section~\ref{sec:hp_tuning}, and show the results in Table~\ref{tab:varying_S_16384}. 



We note an improvement over our previous best performance at $(B,\sigma,S)=(16384, 2.5, 72\mathrm{K})$ referred in Table~\ref{tab:hptuning}. 
Overall, we improved performance by $9\%$ when training from scratch on ImageNet with DP-SGD under $\varepsilon=8$. 
We compare to our reproduction of the previous SOTA of \cite{DeepMindUnlocking} at $30.2\%$ (compared to the results reported in the original paper ($32.4\%$), we still gain $7\%$ of accuracy). 
Thus, we have shown how we can use \BigSigma to perform optimal privacy parameter search while simulating each choice of optimal parameters at a much smaller cost.

\subsection{Ablation}

We now illustrate the benefit of TAN for ablation analysis. 
We study the importance of the order between activation and the normalization layers when training with DP-SGD. 
We also discuss how gathering more training data improves performance while decreasing $\varepsilon$. 
On both experiments, we train a 16-4-WideResnet on CIFAR-10, constant learning rate at $4$, and we are studying $(B_\mathrm{ref},\sigma_\mathrm{ref},S)=(4096,3,2.5\mathrm{k})$.

\paragraph{Pre-activation vs Post-activation Normalization}\label{sec:preac}
\label{preact_norm} Normalization techniques such as BatchNorm \citep{BatchNorm}, GroupNorm (GN) \citep{GroupNorm} or LayerNorm \citep{LayerNorm} help training DNNs. 
Note that BatchNorm is not compatible with DP-SGD because it is not amenable to per-sample gradient computations, we thus resort to GroupNorm.
These normalization layers are usually placed between convolutional layers and activations (e.g., CONV-GN-ReLU). 
\citet{ConvBN} suggest that signal propagation improves when the order is reversed (to CONV-ReLU-GN).

\begin{figure}[t]
    \centering
    \includegraphics[width=\columnwidth]{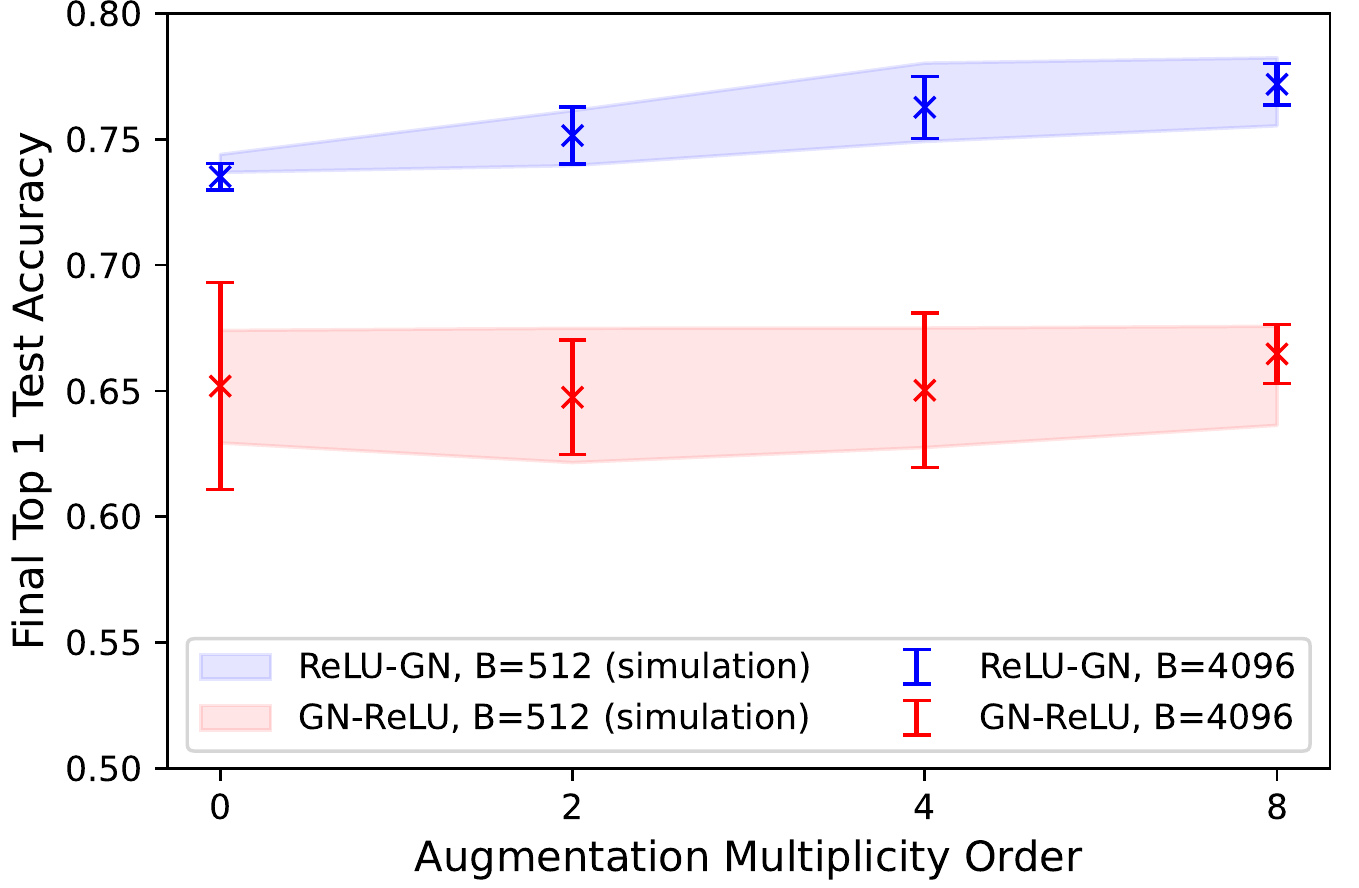}
    \vspace{-2mm}
    \caption{Impact of data augmentation on the test accuracy using pre-activation normalization vs post-activation in a 16-4-WideResnet on CIFAR-10. We compare simulation at $(B,\sigma)=(512,\frac{3}{8})$ and reference $(B_\mathrm{ref},\sigma_\mathrm{ref})=(4096,3)$, both trained for $S=2{,}500$ steps. Confidence intervals are plotted with two standard deviations over 5 runs. Augmentation Multiplicity Order corresponds to the number of augmentations per image, or $K$ in Appendix~\ref{appendix:augmult}. }
    \label{fig:preact_norm}
\end{figure}



We experiment with DP-SGD training using both orders of layers, and display our results in Figure~\ref{fig:preact_norm}. 
We make two observations. First, the reverse order leads to significantly greater performance, and is more robust. Second, the standard order does not benefit from data augmentation.
We observe that the two simulated experiments with $B=512$ represented by lighter colors in Figure~\ref{fig:preact_norm} ($2$ standard deviations around the means) have the same properties. However, each simulation is $8$ times less computationally expensive. Therefore, using \BigSigma through our scaling law can facilitate studying variants of the network architecture while reducing the computational costs.

\paragraph{Quantity of Data}

We now look at how collecting more data affects the tradeoff between privacy and utility. 
We show that doubling the data (from the same distribution) allows better performance with half the privacy budget. 
To this end, we train on portions of the CIFAR-10 training set ($N=50$k) and always report accuracies on the same test set. 
If we multiply by $\beta$ the quantity of data $N_0$ and keep the same $(B,\sigma,S)$, $q$ (and thus $\eta$), is divided by $\beta$  as well. 
We divide $\delta$ by $\beta$ for the accounting. 
We show in Table~\ref{tab:tab_double_data} the effects on $\varepsilon$ and model accuracy.


\begin{table}[!t]
\centering
\vspace*{-4mm}
\caption{
\label{tab:tab_double_data}
Impact of the training set size $N$ on the privacy/utility trade-off. 
We start training on $10\%$ of the data ($N_0=5$K). 
We use $B=4{,}096$, $\sigma=3$ and $S=2{,}500$, with post-activation normalization, and no augmentation. 
Standard deviations are computed over $3$ independent runs.
}
\vspace*{2mm}
\begin{tabular}{*{3}{c}}
\toprule
    \multicolumn{3}{c}{CIFAR-10: $\sigma=3$, $B=4{,}096$, $S=2{,}500$} \\
    \midrule
    $N$ &   $\varepsilon$  &  Test acc (\%)\\
    \midrule
     $5$K &  150.3 & 59.9 $\space (\pm 1)$\\
     $25$K &   13.7 &  71.1 $\space (\pm 0.4)$ \\
     $40$K &   7.3 &  72.9$ \space (\pm 0.1)$ \\
     $50$K &   7.1 &  74.0 $\space (\pm 0.5)$\\
\bottomrule
\end{tabular}
\vspace*{-1mm}
\end{table}





On the one hand, when using $\varepsilon_\mathrm{TAN}$, we can predict the impact on the privacy budget. 
On the other hand, since the global signal-to-noise ratio $N\eta$ is held constant in all experiments, we expect to extract the same amount of information in each setup; adding more data makes this information richer, which explains the gain in accuracy. We show similar results for ImageNet in Appendix~\ref{appendix:more_data}.

%% file: TANIcml2023/conclusion.tex
\section{Conclusion, Limitations and Future Work}

\subsection{Conclusion}

We argue that the total amount of noise (TAN) is a simple but useful guiding principle to experiment with private training.
In particular, we demonstrate that the privacy budget is either a direct function of TAN or can be reduced.
We further show that scaling batch size with noise level using TAN allows for ultra-efficient hyper-parameter search and demonstrate the power of this paradigm by establishing a new state of the art for DP training on ImageNet. 

\subsection{Limitations}
\paragraph{Non-private Hyper-parameter Search.}
We follow the standard practice of not counting hyper-parameter search towards the privacy budget~\citep{li2021large,DPBert}.
Theoretically, each training run should be charged on the overall budget, but in practice it is commonly assumed that the ``bandwidth" of hyper-parameters is too small to incur any observable loss of privacy (see also~\citet{liu2019private} for a theoretically sound way of handling this problem).
If available, one can use a similar public dataset (such as ImageNet) to choose hyper-parameters, and then perform only limited runs on the private dataset.
Finally, we note that training non-private models might not be possible on sensitive data. 
In this case, our hyper-parameter transfer process can not be used.

\subsection{Discussion and Future Work}\label{sec:discussion}
\vspace{0.5em}
\paragraph{Stochasticity in the non Convex Setting}
Varying the batch size at a constant number of steps and a constant $\eta_{\text{step}}$, (and thus constant \BigSigma), we expected a constant test performance.
Indeed, the Gaussian noise in Equation~\ref{eq:DP_SGD} stays the same, the only difference is that the (clipped) gradients are averaged across a different number of samples.
We hypothesize that the better performance at small batch size observed on ImageNet is due to the benefits of stochasticity (i.e., the natural noise of the per-sample gradients).
This is coherent with empirical and theoretical work on the advantages of stochasticity for empirical risk minimization in the non-convex setting~\citep{keskar2016large,masters2018revisiting,
pesme2021implicit}. 

In particular, it is consistent with the (non-private) empirical work of \citet{SmallBS}, which observe that for a fixed number of steps, small batches perform better than large batches when training DNNs. 


\vspace{0.5em}
\paragraph{Theoretical Analysis of \BigSigma in the Convex Setting}
Convergence theory has been thoroughly studied for DP-SGD in convex, strongly convex, and nonconvex (stationary point convergence) settings \citep{bassily2014private, wang2017differentially, feldman2018privacy}. 
For example, under the convex assumption, the excess bound given in Theorem 2.4 of \citet{bassily2014private} with decreasing learning rate can be extended to mini batch training, and does not change when we hold $S$ and $\eta_{\text{step}}$ constant for different batch sizes.
The same observation holds for a constant learning rate, which means that the optimal learning rate (with respect to this bound) is the same for all batch sizes with our scaling strategy, which is what we observe in practice (Table \ref{tab:hptuning}). 

However, if we model the natural noise of the gradients for SGD, the upper bound will have an additional dependency on the batch size \citep{gower2019sgd}, which could be informative for our scaling laws. 
We defer investigation of this assumption to future work.

\paragraph{Better Accounting.}
We believe that the important increase in the privacy budget $\varepsilon$ as the noise level $\sigma$ decreases is a real phenomenon and not an artifact of the analysis.
Indeed, DP assumes that the adversary has access to all model updates, as is the case for example in FL.
In such cases, a noise level that is too low is insufficient to hide the presence of individual data points and makes it impossible to obtain reasonable privacy guarantees.
In the centralized case however, the adversary does not see intermediate models but only the final result of training. 
Some works have successfully taken into account this ``privacy amplification by iteration" idea~\citep{feldman2018privacy,ye2022differentially} but results are so far limited to convex models.



\newpage

%% file: appendix.tex
\newpage 

\section{More data: ImageNet}\label{appendix:more_data}

We show in Table~\ref{tab:tab_double_data_ImageNet} that similarly to the experiments in CIFAR-10, doubling the training data on ImageNet improves the accuracy while diving $\varepsilon$ by $2$.
We also demonstrate that our scaling strategy can accurately detect the gain of accuracy. We compare training on half of the ImageNet training set ($N=600k$) and the entire training set ($N=1.2M$). 

\begin{table}[b]
    \centering
    \caption{Impact of adding more data on ImageNet. The ``Simulated Gain'' column corresponds to the accuracy gain we observe when simulating at lower compute using our scaling strategy for $B=256$. The ``Gain'' column corresponds to the real gain at $B=16384$. }
    \vspace*{2mm}
    \label{tab:tab_double_data_ImageNet}
\begin{tabular}{*{6}{c}}
\toprule
    \multicolumn{6}{c}{Imagenet: $\sigma_{ref}=2.5$, $B_{ref}=16384$, $S=72k$} \\
    \midrule
    N &  $\delta$ & $\varepsilon$  & $\varepsilon_{TAN}$ & Gain & Simulated Gain\\
    \midrule
     0.6M & $16.10^{-7}$ & 17.98 & 18.06 & / & / \\
     1.2M &  $8.10^{-7}$  & 8.00 & 8.26 & +1.3\% & +1.5\% \\
\bottomrule
\end{tabular}
\end{table}

\section{Choice of architecture and optimizer}\label{appendix:architecture}


In this section, we give more details about our choice of architecture and optimizer on ImageNet. In particular, we noticed that DP-SGD without momentum is always optimal, even with ViTs, and that NF-ResNets-50 performed the best. 

\paragraph{Architecture.}When training with DP-SGD, the goal is to find the best possible local minimum within a constrained number of steps $S$, and with noisy gradients. However, architectures and optimizers have been developed to ultimately achieve the best possible final accuracy with normal updates. To illustrate this extremely, we train a Vision Transformer (ViT) \citep{ViT} from scratch on ImageNet using DP-SGD. \citet{deit} have succeeded in achieving SOTA performance in the non-private setting, but with a number of training steps higher than convolution-based architectures. A common explanation is that ViTs have less inductive bias than CNNs: they have to learn them first, and that can be even harder with noisy gradients. And if they are successful, they have lost the budget for gradient steps to learn general properties of images.

\begin{figure}[t!]
    \centering
    \vspace{-5mm}
    \includegraphics[width=0.9\textwidth]{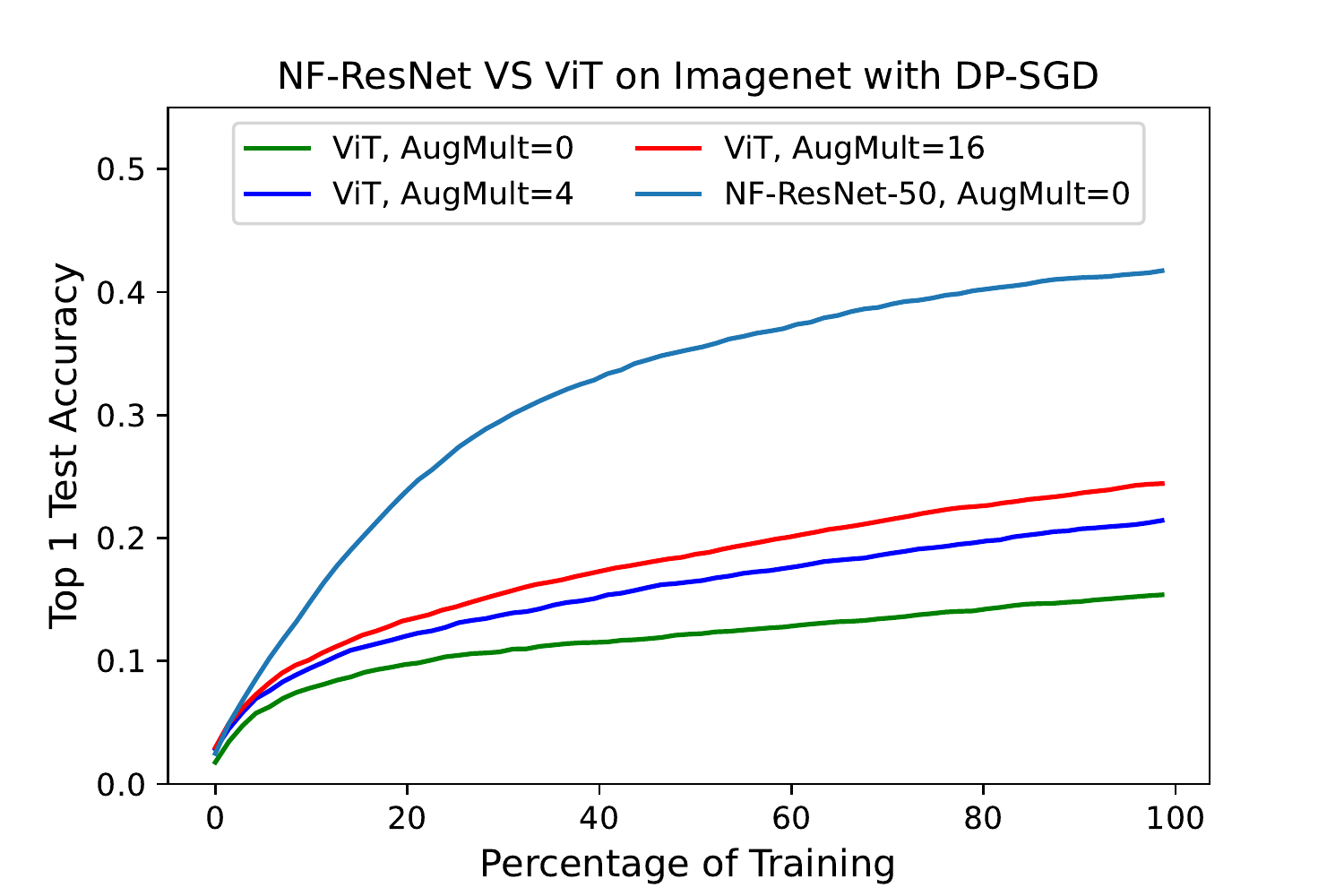}
    \caption{Training a ViT from scratch on ImageNet with DP-SGD. We simulate training with our scaling strategy and $B=256$. We observe that the accuracies are not as good as for  NF-ResNets, and that Augmentation Multiplicity plays a more important role.}
    \label{fig:ViT}
\end{figure}

We used our scaling strategy (keeping $\eta_{step}$ and $S$ constant) to simulate the DP training with different architectures at low compute, studying noisy training without the burden of DP accounting. The best simulated results were obtained with a NFResNet-50 \citep{NFnet} designed to be fast learners in terms of number of FLOPS. The worst results were obtained with ViTs, and intermediate results with classical ResNets. In Figure ~\ref{fig:ViT}, we compare different training trajectories of a ViT and a NF-ResNet.

\paragraph{Optimizer}Using our simulation scheme, we found that DP-SGD with no momentum and a constant learning rate is the best choice for all architectures. We also tried DP-Adam, DP-AdamW with a wide range of parameters. It is surprising to find that this is the case for ViTs, as without noisy, the Adam type optimizers perform better \citep{deit}. This highlights the fact that training with DP-SGD is a different paradigm that requires its own tools.

Using \BigSigma allowed us to explore and compare different architectures and optimizers, which would have been computationally impossible in the normal DP training setting at $B=16384$.

\section{Augmentation Multiplicity}\label{appendix:augmult}

\paragraph{Augmentation Multiplicity}(AugMult) was introduced by \cite{DeepMindUnlocking} in the context of DP-SGD. The authors average the gradients of different augmentations of the same image before clipping the per-sample gradients, using the following formula (where $\zeta$ is a standard Gaussian variable):

\begin{align} 
\label{eq:AugMult}
w^{t+1}=w^{t+1} - \eta_{t}\left( \frac{1}{B}\sum_{i \in B_t} \frac{1}{C}\text{clip}_C\left(\frac{1}{K}\sum_{j \in K_t} \nabla_j(w^{(t)})\right) + N\left(0,\frac{\sigma^2}{B^2}\right)\right)
\end{align}

\begin{figure}[h!]
    \centering
    \includegraphics[width=0.46\textwidth,height=4.5cm]{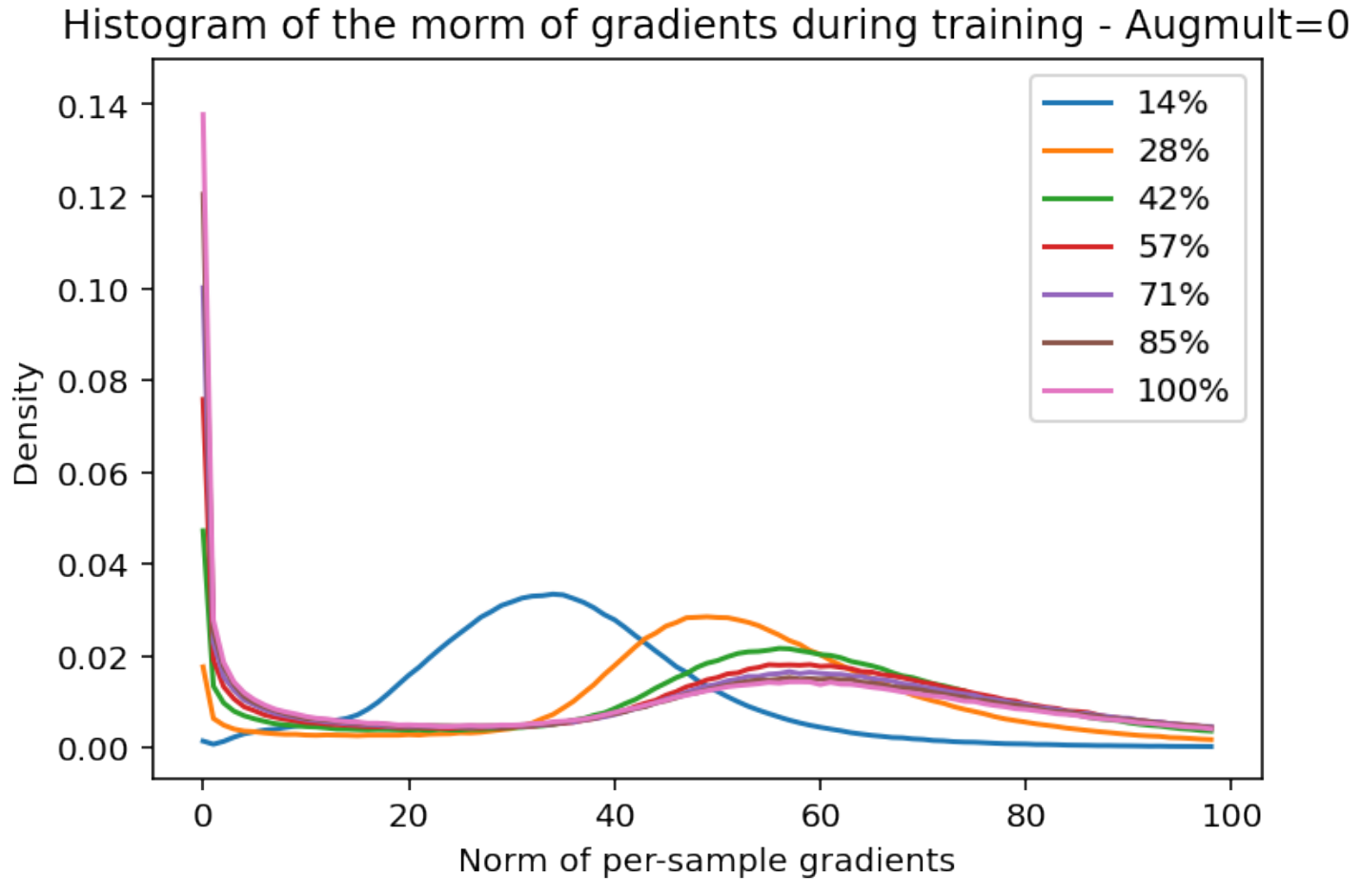}
    \includegraphics[width=0.50\textwidth,height=4.5cm]{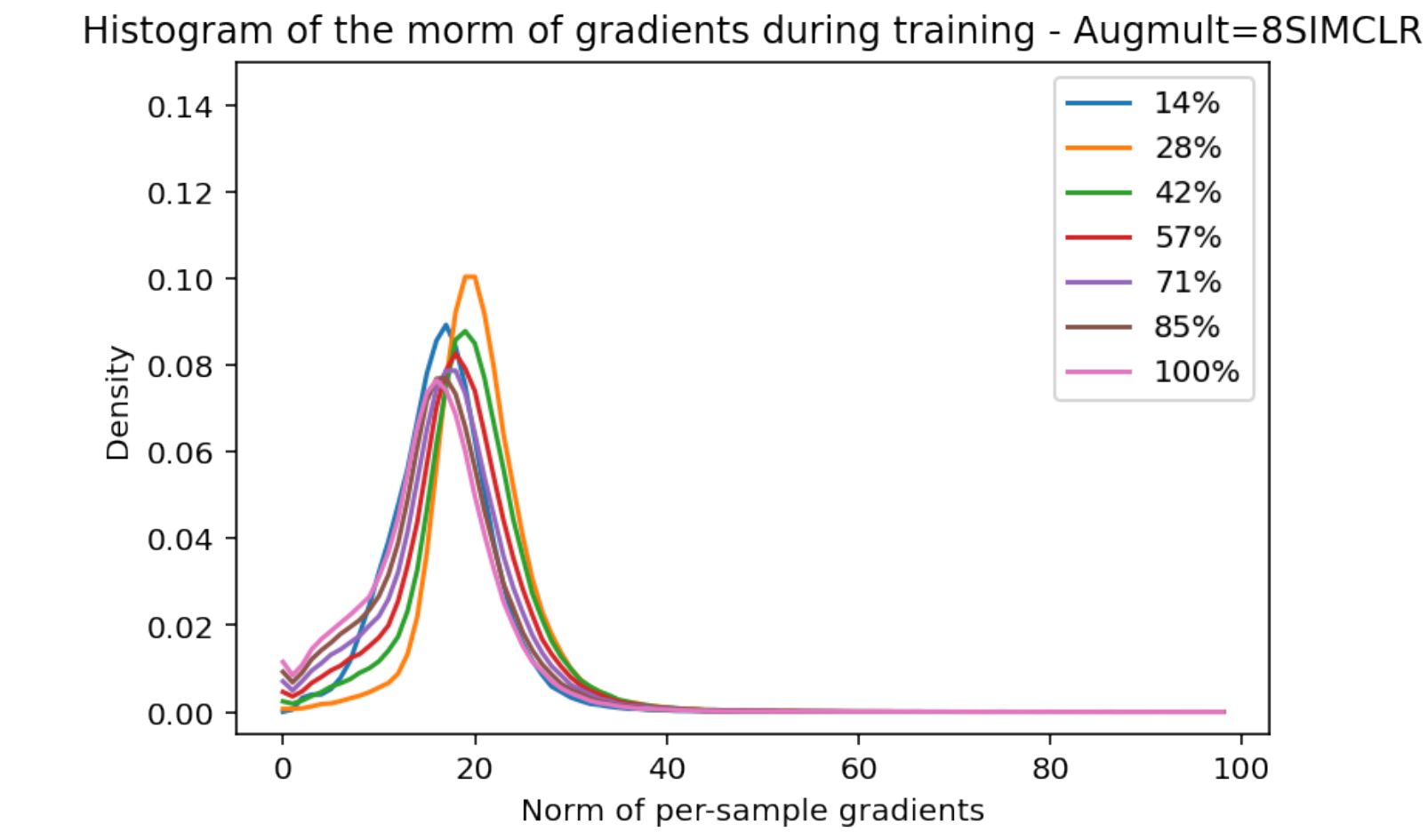}
    \caption{Histograms of the norms of the \textbf{average across all augmentations for each image} of the per-sample gradients, before clipping and adding noise. On the left, we see that without augmentation, an increasing number of examples have their gradients going to zero during training. On the right, we see that when using a strong augmentation technique (SimCLR, \citep{simclr}), the gradients are more concentrated during all the training.}
    \label{histograms_gradients_augmult}
\end{figure}

Compute scales linearly with the AugMult order $K$. Our intuition on the benefits of AugMult is that difficult examples (or examples that fall out of the distribution) become easier when using this augmentation technique. On the other hand, without AugMult, simple examples are learned to be classified early in training, resulting in a gradient close to $0$ when used without augmentation. Because we are training for a fixed number of steps, it is a waste of gradient steps (i.e. privacy budget). With AugMult, the network may still be able to learn from these examples.  Figure \ref{histograms_gradients_augmult} shows the histograms of the norms of the \textbf{average over all augmentations for each image} of the per-sample gradients, before clipping and adding noise in \eqref{eq:AugMult} at different times of training. \\

